\pdfoutput=1
\documentclass[11pt]{article}

\usepackage{acl}
\usepackage{amsmath, amsfonts, amssymb, amsthm, xspace, color}
\usepackage{booktabs} % For formal tables

\usepackage{subfigure, multirow, tabularx} % For subfigure
\usepackage{enumitem}
\usepackage{flushend}
\usepackage[T1]{fontenc}
\usepackage{epstopdf}
\usepackage{balance}
\usepackage{graphicx}
\usepackage[noend,ruled,linesnumbered]{algorithm2e} % For algorithm2e
\usepackage{url}
\usepackage{wrapfig,lipsum}
\usepackage{makecell}
\usepackage{wrapfig}

\usepackage{listings}
\lstdefinestyle{mystyle}{
  basicstyle=\ttfamily,
  frame=single,
  breaklines=true,
  breakindent=1pt,
  backgroundcolor=\color{blue!5}, % Change the background color
}

\newcommand{\hide}[1]{} %hide

\newcommand{\nop}[1]{}
\newcommand{\mquote}[1]{{``\emph{#1}''}}

\newtheorem{thm:def}{Definition}
\newtheorem{thm:eg}{Example}
\newtheorem{thm:lem}{Lemma}
\newtheorem{thm:obs}{Observation}
\newtheorem{thm:req}{Requirement}
\newtheorem{thm:prop}{Proposition}
\newtheorem{thm:principle}{Principle}
\newtheorem{thm:thm}{Theorem}
\newtheorem{thm:corollary}{Corollary}

\usepackage{bbm}
\usepackage{adjustbox}
\usepackage[super]{nth}

\usepackage{times}
\usepackage{latexsym}

\usepackage[utf8]{inputenc} % allow utf-8 input
\usepackage[T1]{fontenc}    % use 8-bit T1 fonts
\usepackage{hyperref}       % hyperlinks
\usepackage{url}            % simple URL typesetting
\usepackage{booktabs}       % professional-quality tables
\usepackage{amsfonts}       % blackboard math symbols
\usepackage{nicefrac}       % compact symbols for 1/2, etc.
\usepackage{microtype}      % microtypography
\usepackage{xcolor}         % colors
\usepackage{enumitem}

\usepackage{multirow}
\usepackage{graphicx}
\usepackage{subfigure}
\usepackage{pgfplots}
\pgfplotsset{compat=1.7}
\usepgfplotslibrary{groupplots}
\definecolor{g900grey}{HTML}{202124}
\definecolor{g900blue}{HTML}{174EA6}
\definecolor{g900red}{HTML}{A50E0E}
\definecolor{g900yellow}{HTML}{E37400}
\definecolor{g900green}{HTML}{0D652D}
\definecolor{g700grey}{HTML}{5F6368}
\definecolor{g700blue}{HTML}{1967d2}
\definecolor{g700red}{HTML}{c5221f}
\definecolor{g700yellow}{HTML}{f29900}
\definecolor{g700green}{HTML}{188038}
\definecolor{g500grey}{HTML}{9AA0A6}
\definecolor{g500blue}{HTML}{4285F4}
\definecolor{g500red}{HTML}{EA4335}
\definecolor{g500yellow}{HTML}{FBBC04}
\definecolor{g500green}{HTML}{34A853}
\definecolor{g400blue}{HTML}{669DF6}
\definecolor{g300green}{HTML}{81C995}
\definecolor{g300grey}{HTML}{DADCE0}
\definecolor{g300blue}{HTML}{8AB4F8}
\definecolor{g200green}{HTML}{A8DAB5}
\definecolor{g200blue}{HTML}{AECBFA}
\definecolor{g100grey}{HTML}{F1F3F4}
\definecolor{g100green}{HTML}{CEEAD6}
\definecolor{g100blue}{HTML}{D2E3FC}
\definecolor{g050green}{HTML}{E6f4EA}
\definecolor{g050blue}{HTML}{E8F0FE}
\definecolor{g100red}{HTML}{FAD2CF}
\definecolor{g100yellow}{HTML}{FEEFC3}
\definecolor{g100green}{HTML}{CEEAD6}

\definecolor{g700grey}{HTML}{5F6368}
\definecolor{g700blue}{HTML}{1967d2}
\definecolor{g700red}{HTML}{c5221f}
\definecolor{g700yellow}{HTML}{f29900}
\definecolor{g700green}{HTML}{188038}

\usepackage{amsmath}
\usepackage{amssymb}
\usepackage{mathtools}
\usepackage{amsthm}
\usepackage{footmisc}

\newcommand{\commentout}[1]{}

\title{Multilingual Fine-Grained News Headline Hallucination Detection}

\author{Jiaming Shen$^\dagger$ \hspace*{1cm} Tianqi Liu$^\dagger$ \hspace*{1cm}  Jialu Liu$^\dagger$ \hspace*{1cm} Zhen Qin$^\dagger$ \\
\bf{Jay Pavagadhi$^\ddagger$ \hspace*{1cm} Simon Baumgartner$^\dagger$ \hspace*{1cm} Michael Bendersky$^\dagger$} \\
$^\dagger$ Google Research \quad
$^{\ddagger}$ Google \quad \\
\small{\texttt{\{jmshen, tianqiliu, jialu, zhenqin, jaynp, simonba, bemike\}@google.com}}
}

\begin{document}
\maketitle
%!TEX root = main.tex
% UTF-8 encoding
\begin{abstract}

The popularity of automated news headline generation has surged with advancements in pre-trained language models.
However, these models often suffer from the ``hallucination'' problem, where the generated headline is not fully supported by its source article.
Efforts to address this issue have predominantly focused on English, using over-simplistic classification schemes that overlook nuanced hallucination types.
In this study, we introduce the first multilingual, fine-grained news headline hallucination detection dataset that contains over 11 thousand $\langle$article, headline$\rangle$ pairs in 5 languages, each annotated with detailed hallucination types by experts.
We conduct extensive experiments on this dataset under two settings.
First, we implement several supervised fine-tuning approaches as preparatory solutions and demonstrate this dataset's challenges and utilities.
Second, we test various large language models' in-context learning abilities and propose two novel techniques, language-dependent demonstration selection and coarse-to-fine prompting, to boost the few-shot hallucination detection performance in terms of the example-F1 metric.
We release this dataset to foster more research in multilingual, fine-grained headline hallucination detection.

\end{abstract}

%!TEX root = main.tex
% UTF-8 encoding

\section{Introduction}\label{sec:intro}

% Figure 1: demo example in intro section
\begin{figure}[!t]
    \centering
    \includegraphics[width=0.98\linewidth]{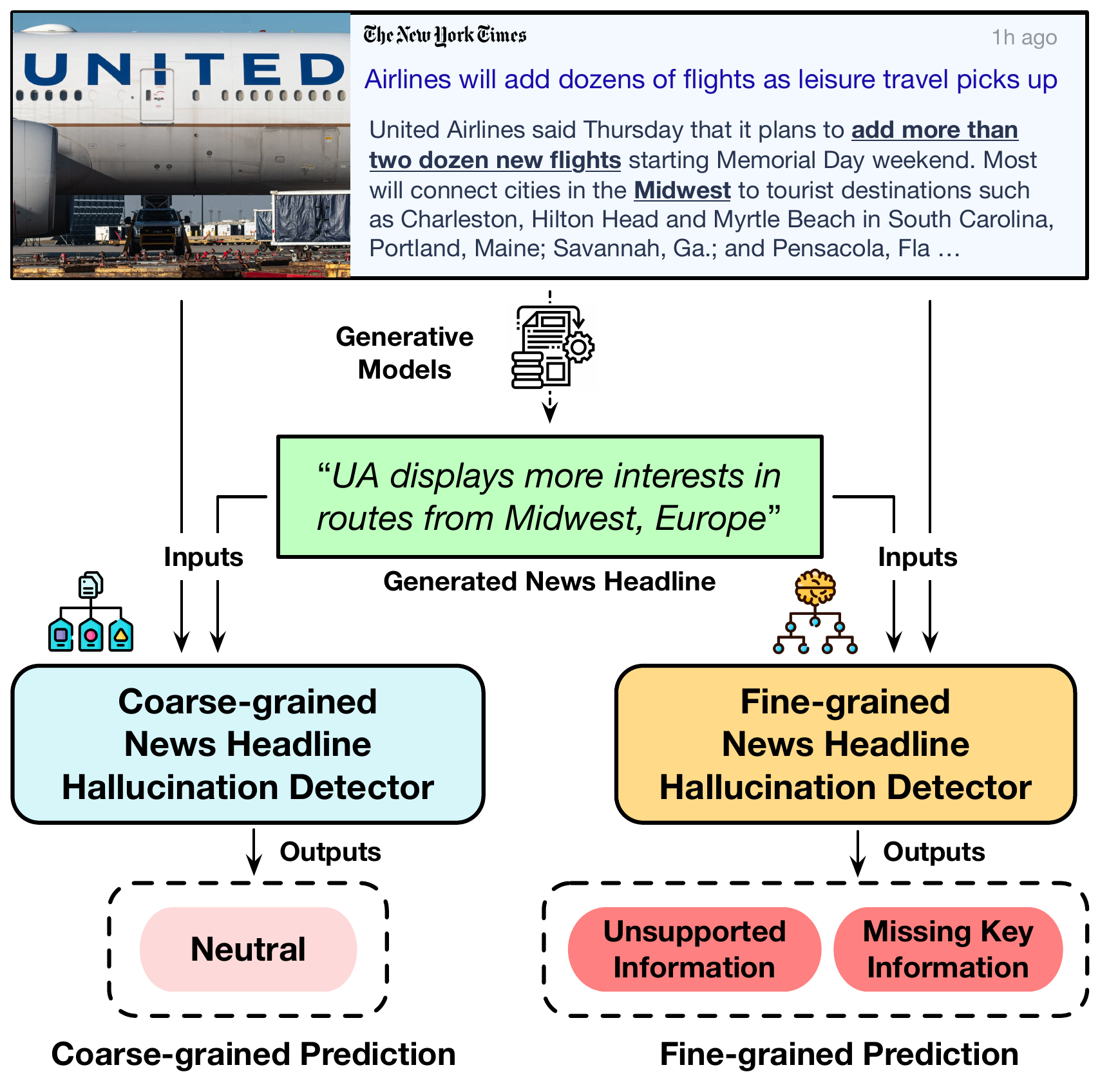}
    \vspace{-0.3ex}
    \caption{A comparative example of headline hallucination detection at different levels of granularity. The fine-grained hallucination detector goes beyond traditional 3-class label \textbf{Neutral} and  offers more nuanced predictions like \textbf{Unsupported Information} (because the article does not references ``Europe'') and \textbf{Missing Key Information} (as the headline omits the crucial detail that ``new routes in the Midwest are being added''). 
    }
    \label{fig:intro}
\end{figure}

A news headline provides a concise summary of its corresponding news article, enabling readers to quickly grasp the essence of a news story.
Numerous generative models~\cite{Gu2020GeneratingRH, Cai2023GeneratingUN, Ding2023HarnessingTP} have been developed to automate the process of condensing a news article into its headline, achieving generally commendable quality.
However, people note that these models often encounter the hallucination problem, where the produced headline does not fully align with the source article's content.
For example, as shown in Figure~\ref{fig:intro}, the generation model is given an article about ``UA adds new routes in Midwest" and outputs the headline \mquote{UA displays more interests in routes from Midwest, Europe}.
This headline is considered as a hallucination because it references "Europe" which is absent from the article, and omits the crucial detail that "new routes in the Midwest are being added".

To mitigate these hallucinations, various studies propose to pre-process the training corpus of generation models by removing or re-weighting possibly hallucinated examples~\cite{Nan2021EntitylevelFC, Aharoni2022mFACEMS, Qiu2023DetectingAM}.
Another line of work proposes to first detect hallucinations in generated outputs and then filter them in a post-processing stage~\cite{Honovich2022TRUERF, Shen2023WhyIT}.
These approaches typically adopt a binary or three-way classification scheme and focus on examples written exclusively in English or translated from a single English source.
Despite some promising results, it remains unclear how these approaches can capture more fine-grained hallucination error types in multilingual news articles and inform more nuanced decision making process.

In this work, we propose a new task --- \emph{fine-grained headline hallucination detection} and study it in the multilingual setting.
This objective is to identify a \emph{set} of fine-grained entailment relations between a given news article and its headline.
Taking the example in Figure~\ref{fig:intro} for instance, we aim to advance beyond a simple ``Neutral'' classification and provide more fine-grained predictions: (1) ``Unsupported Information'' due to the article's lack of references to "routes from Europe", and (2) ``Missing Key Information'' because the headline fails to capture a core message of the article --- the introduction of new routes in the Midwest.

To advance research in this area, we introduce the first \underline{M}ultilingual \underline{F}ine-grained \underline{H}eadline \underline{H}allucination \underline{D}etection (MFHHD) dataset, featuring 11,469 examples across 5 languages.
Each example comprises a news article, a generated news headline, a coarse-grained hallucination label, and a set of fine-grained hallucination labels annotated by 2 to 4 dedicated domain experts fluent in the language of the original article.
Additionally, for examples labeled as ``Neutral'' or ``Contradict'', annotators will provide a natural language justification for their fine-grained annotations. 

The introduction of this new MFHHD dataset presents intriguing research challenges, such as identifying complex, nuanced types of hallucination errors and exploring whether existing hallucination detection moethods, previously English-centric, can be adapted for multilingual use.
To answer these questions, we carry out extensive experiments in both supervised fine-tuning and few-shot learning scenarios. 
In the supervised fine-tuning context, we find that model pre-training on natural language inference datasets and incorporation of natural language explanations into seq2seq based classifier can both enhance the detection performance.
In the few-shot learning domain, we evaluate various large language models (e.g., ChatGPT~\cite{chatgpt}, PaLM2 variants~\cite{anil2023palm}) and observe that they perform worse than the smaller fine-tuned models (e.g., mT5-XXL~\cite{Xue2020mT5AM}).
To improve these LLMs' in-context learning capabilities, we introduce two prompting techniques: (1) \emph{language-dependent demonstration selection} which dynamically chooses few-shot examples in the same language as the test query example, and (2) \emph{coarse-to-fine prompting} that guides LLMs to generate a coarse-grained prediction before making fine-grained hallucination type predictions.
Both techniques can significantly enhance LLM's few-shot learning effectiveness and boost the detection performance in terms of the example-F1 metric.

\smallskip
\noindent \textbf{Contributions}.
The major contributions of this paper are summarized as follows:
(1) We introduce a novel task, fine-grained headline hallucination detection, aimed at identifying more nuanced hallucination error types;
(2) We create a new multilingual fine-grained hallucination detection dataset MFHHD, curated by news domain experts; and
(3) We conduct extensive experiments on the MFHHD dataset, delving into its complexities and offering valuable insights for improving the accuracy of fine-grained hallucination detection in both supervised and few-shot learning settings.

%!TEX root = main.tex
% UTF-8 encoding

\section{Problem Formulation}\label{sec:problem}

In this study, we represent both a news \emph{article} $d$ and a news \emph{headline} $h$ as a sequence of tokens.
The ideal purpose of a news headline is to provide a concise summary of the news article.
However, when automatically generated, the headline can hallucinate and misrepresent the original intent of the source article.
The \textbf{coarse-grained headline hallucination detection} task inputs a pair of news article and headline $\langle d_i, h_i \rangle$ and outputs its coarse-grained entailment relation $l_i$ indicating whether the headline is fully supported, directly contradicts, or remains neutral with respect to the article.

In many real-world applications, we notice this the three-way entailment classification schema is too coarse-grained and fails to pinpoint the exact hallucination reasons (e.g., the headline includes an incorrect number or reports a person's subjective opinion as a fact).
Therefore, we propose the \textbf{fine-grained headline hallucination detection} task that returns \emph{a set of} fine-grained entailment relations $\mathcal{R}_i = \{R_i^{1}, R_i^{2}, \dots\}$ for a pair of input article and headline $\langle d_i, h_i \rangle$.
Each relation $R_i^{j}$ specifies a detailed reason why the given headline $h_i$ either supports, contradicts, or is neutral in relation to the corresponding article $d_i$.

%!TEX root = main.tex
% UTF-8 encoding

\section{MFHHD Dataset}\label{sec:dataset_construction}

In this work, we collect the first \underline{M}ultilingual \underline{F}ine-grained \underline{H}eadline \underline{H}allucination \underline{D}etection (MFHHD) dataset that contains 11,469 examples across 5 languages (English, Spanish, German, French, Portuguese).
Each example includes a news article, a news headline, a coarse-grained hallucination label, a set of fine-grained hallucination labels, and an optional set of natural language annotation justifications.
The dataset is currently available at: \url{https://bit.ly/MFHHD-dataset}.

\subsection{Dataset Construction}
We follow the same procedure as in \citep{Shen2023WhyIT} to sample a set of news articles along with their headlines generated from NHNet~\cite{Gu2020GeneratingRH}\footnote{\small More results are discussed in Appendix~\ref{app:label_source}.}. 
We examine these sampled $\langle$article, headline$\rangle$ pairs and discuss with multiple news domain experts to outline the following 7 fine-grained hallucination types (see Figure~\ref{fig:demo_examples} in Appendix~\ref{app:demo_examples} for examples of each hallucination type).

\begin{figure*}[!t]
    \centering
    % \vspace{-ex}
    \includegraphics[width=0.98\linewidth]{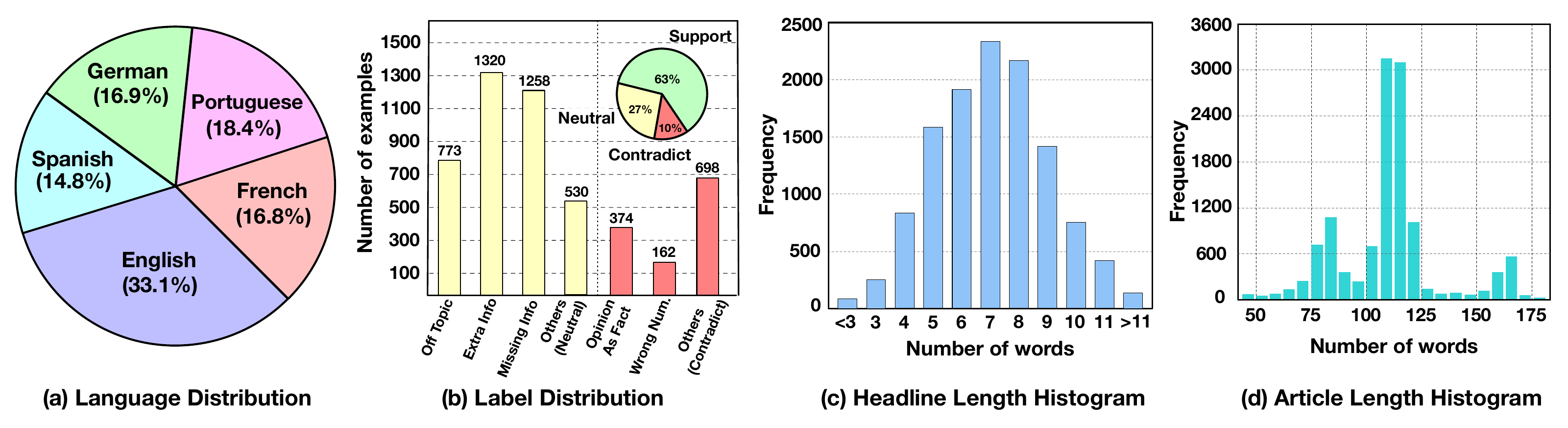}
    \vspace{-0.3ex}
    \caption{Analysis of our multilingual fine-grained headline hallucination detection (MFHHD) dataset.}
    % \vspace{-2ex}
    \label{fig:data_analysis}
\end{figure*}

\noindent (1) \textbf{Neutral (extra info)}: the headline contains unsupported additional information that cannot be verified by the given article.

\noindent (2) \textbf{Neutral (missing info)}: the headline misses important information (e.g., key dates, locators) and thus changes the scope/emphasis of the article. 

\noindent (3) \textbf{Neutral (off topic)}: the headline and article discuss two completely different topics.

\noindent (4) \textbf{Neutral (others)}: the catch-all option for all other forms of headlines that neither fully supports nor directly contradicts the article.

\noindent (5) \textbf{Contradictory (opinion as fact)}: the article states an opinion or unconfirmed rumor while the headline interprets it as a factual statement.

\noindent (6) \textbf{Contradictory (wrong number)}: the headline includes an incorrect important number that directly contradicts the news article.

\noindent (7) \textbf{Contradictory (others)}: the catch-all option for all other forms of direct contradictions between the headline and the news article.

We prepare a detailed curation guideline that lists the definitions and representative examples for the above 7 fine-grained hallucination types plus 1 non-hallucination type (i.e., ``Support'') and train all the annotators for two rounds.
All annotators are full-time journalist degree holders and speak the same language of the annotated article.
Due to some policy constraints, we cannot report their detailed compensation here but we guarantee their pay is definitely above the corresponding local minimum wage.
Given a pair of article and headline, they are instructed to first choose one coarse-grained type (``Support'', ``Neutral'', or ``Contradict'') and then to select all fine-grained hallucination types of this example.
Additionally, if they label one example as ``Neutral'' or ``Contradict'', we encourage them to provide an additional natural language explanation to justify their fine-grained annotations.
Each example undergoes initial evaluation by two annotators and if they disagree with each other at the coarse-grained label, we engage two additional curators to thoroughly review the example.
Finally, we retain all examples that receive a majority consensus at the coarse-grained label level and preserve all corresponding fine-grained labels associated with these chosen coarse-grained labels.
The initial round of annotator agreement (at the coarse-grained level) is 74.3\%.
For the remaining 25.7\% of examples that have two rounds of annotations from 4 raters, their inter-rater agreement is 0.6642 Cohen's Kappa and thus can be considered as substantial agreement.

\subsection{Dataset Analysis}
We analyze some properties of our MFHHD dataset and show the results in Figure~\ref{fig:data_analysis}.

First, we can see that over 65\% of examples in our dataset are non-English examples and their corresponding languages are evenly distributed in German, Spanish, French, and Portuguese.

Second, we analyze the coarse-grained label distribution and observe about one-third of examples are marked as ``Neutral'' or ``Contradict''. 
Furthermore, each ``Neutral'' example has an average 1.3 fine-grained labels and about 29\% of examples have more than one fine-grained label.
Conversely, only 4.1\% of ``Contradict'' examples have more than one fine-grained label. One possible explanation is that raters tend to give a single most severe contradictory reason instead of selecting multiple fine-grained ones.
In a holistic view, we can see about 21\% of all hallucinated examples have more than one fine-grained label, which necessitates our formulation of fine-grained hallucination detection task as a multi-label classification problem.

Finally, we analyze the textual information in our MFHHD dataset.
We draw the length histograms of news articles and headlines in Figure~\ref{fig:data_analysis}(c)(d) where we can see the median headline and article length are 7 and 106 words, respectively. 
Furthermore, over 96\% of ``Neutral'' or ``Contradict'' examples have at least one natural language explanation and about half of them have two or more explanations.
These human written explanations have 18 words in median and provide useful signals for detecting headline hallucinations. 

%!TEX root = main.tex
% UTF-8 encoding

\section{Supervised Fine-grained Headline Hallucination Detection}\label{sec:supervised_experiments}

In this section, we experiment a set of supervised methods for multilingual fine-grained headline hallucination detection.
The goal is to better understand the characteristics and challenges of our MFHHD dataset and to share some valuable insights that can help later LLM-based few-shot method designs (c.f. Section~\ref{sec:few-shot-exp}).

\subsection{Experiment Settings}\label{subsec:sft_exp_settings}

\noindent \textbf{Dataset.}
Given the curated MFHHD dataset, we first create the test set by randomly selecting 1000 English examples and 500 examples for each of the remaining languages (German, French, Spanish, Portuguese).
Then, we use the remaining 8,469 examples for training models and test their performances on the above selected 3,000 test examples.

\begin{figure}[!t]
    \centering
    % \vspace{-ex}
    \includegraphics[width=0.98\linewidth]{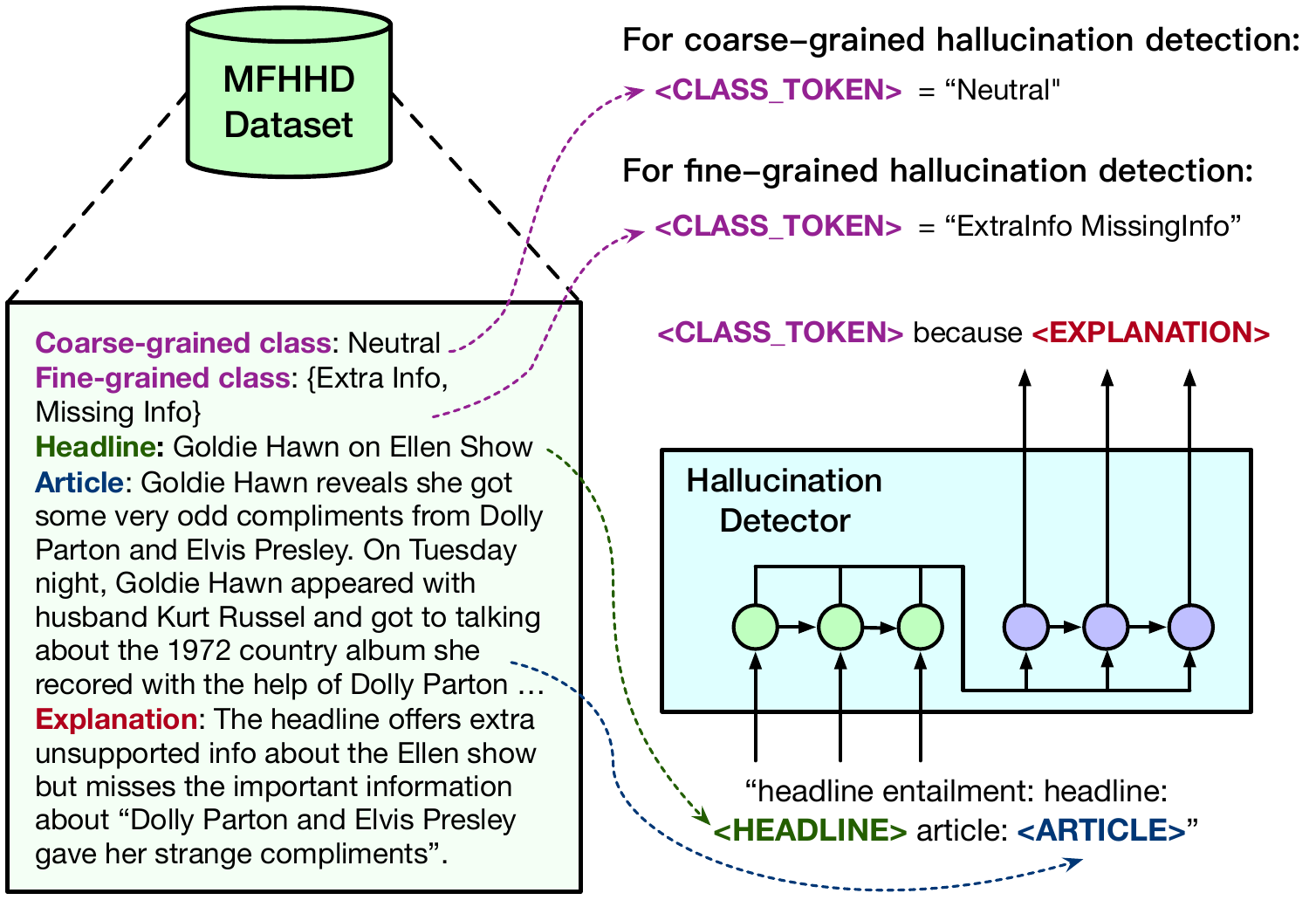}
    \vspace{-0.3ex}
    \caption{Detecting news headline hallucinations with models of the encoder-decoder architecture.
    }
    \vspace{-2ex}
    \label{fig:sft_seq2se_model}
\end{figure}

\smallskip
\noindent \textbf{Compared Methods.}
We compare the following representative methods for the multilingual headline hallucination detection task:
\begin{itemize}[leftmargin=*, itemsep=-1pt]
    \item mDeBERTa$_{{base}}$: The multilingual version of DeBERTa~\cite{He2023DeBERTaV3ID} which enhances the original BERT model~\cite{Devlin2019BERTPO} using replaced token detection as the pretraining task and pre-trained on CC100 multilingual data. We concatenate the headline and the article text (with a special separator token [SEP]) and feed it into the mDeBERTa$_{{base}}$ model for prediction.
    \item mDeBERTa$_{{base}}$ + NLI: We first adopt the above mDeBERTa$_{{base}}$ model and further pre-train it on various natural language inference (NLI) datasets including XNLI~\cite{Conneau2018XNLIEC} and the translated version of MNLI~\cite{MNLI}, ANLI~\cite{nie2019adversarial} and WANLI~\cite{Liu2022WANLIWA}. Then, we fine-tune the model on our MFHHD dataset.
    \item mT5$_{{xxl}}$: The multilingual version of T5~\cite{Xue2020mT5AM}, an encoder-decoder model with strong representation power. We input the concatenated headline and article into the encoder and requires the decoder to output a single token indicating the final predicted coarse-grained class (or a sequence of tokens, each represents one fine-grained hallucination class).
    \item mT5$_{{xxl}}$ + Exp: We incorporate human written natural language explanations into the mT5$_{{xxl}}$ model by requiring its decoder to output the class token(s) followed by the explanation. See Figure~\ref{fig:sft_seq2se_model} for a reference.
    \item mT5$_{{xxl}}$ + NLI: Similar to mDeBERTa$_{{base}}$ + NLI, we pre-train the mT5$_{{xxl}}$ model on NLI datasets and fine-tune it on MFHHD dataset.
    \item mT5$_{{xxl}}$ + NLI + Exp: The combination of mT5$_{{xxl}}$ + Exp and mT5$_{{xxl}}$ + NLI where we incorporate explanations information during the MFHHD fine-tuning stage.
\end{itemize}
For the last four encoder-decoder based models, we evaluate their abilities to detect both coarse-grained and fine-grained hallucinations.
We map fine-grained predictions into their corresponding coarse-grained hallucination labels.
For example, the fine-grained model output ``Off-Topic Missing-Info'' is mapped to the ``Neutral'' class.\footnote{\small Although the model can in theory output multiple incompatible fine-grained class tokens (e.g. ``Off-Topic Incorrect-Number'' where one token corresponds to the ``Neutral'' class while the other belongs to the ``Contradict'' class), we do not witness such a case in practice for the supervised setting.}

We use Huggingface library~\cite{Wolf2020HuggingFacesTS} to implement mDeBERTa$_{{base}}$ and mDeBERTa$_{{base}}$ + NLI.
For the remaining four mT5-based models, we develop them based on the T5X library\footnote{\scriptsize \url{https://github.com/google-research/t5x}}\cite{roberts2022t5x}.
Appendix~\ref{app:sft_exp_details} provides more implementation details and hyper-parameter settings.

\smallskip
\noindent \textbf{Evaluation Metrics.}
We evaluate coarse-grained hallucination detection performance using standard multi-class classification metrics including ``Accuracy'' and ``Weighted-F1''.
The Weighted-F1 considers the number of true instance for each class and thus account for class imbalance.
For fine-grained hallucination detection, we formulate it as a multi-label classification problem and follow previous studies~\cite{prabhu2018parabel,Shen2021TaxoClassHM} to use the ``Example-F1'' metric for evaluation.
% The Example-F1 calculates the average F1 scores for each test $\langle$document, headline$\rangle$ pair as follows:
The Example-F1 is calculated as follows:
\begin{equation*}
\small
    \text{Example-F1} = \frac{1}{N} \sum_{i=1}^{N} \frac{2|\mathcal{R}_{i}^{gt} \cap \mathcal{R}_{i}^{pred}|}{ |\mathcal{R}_{i}^{gt}| + |\mathcal{R}_{i}^{pred}| },
\end{equation*}
where $N$ is the number of test examples; $\mathcal{R}_{i}^{gt}$ and $\mathcal{R}_{i}^{pred}$ stand for the ground truth and model predicted fine-grained hallucination class set of test example $\langle d_i, h_i \rangle$, respectively.

% Figure: seq2seq based classifier for illustration

%% Table: Supervised Results
\begin{table}[!t]
\centering
\scalebox{0.68}{
\begin{tabular}{lcc|c}
\toprule
\bf Methods & \bf Accuracy & \bf Weighted-F1 & \bf Example-F1  \\
\midrule
\multicolumn{2}{l}{\bf Coarse-grained Detection} & & \\
\quad mDeBERTa$_{base}$ & 63.70 & 49.57 & --- \\
\quad mDeBERTa$_{base}$ + \text{NLI} & 66.80 & 62.30 & --- \\
\quad mT5$_{xxl}$ & 71.20 & 69.68 & --- \\
\quad mT5$_{xxl}$ + \text{Exp} & 73.23 & 71.82 & --- \\
\quad mT5$_{xxl}$ + \text{NLI} & 72.60 & 71.52 & --- \\
\quad mT5$_{xxl}$ + \text{NLI} + \text{Exp} & \bf 73.97 & \bf 73.11 & --- \\
\midrule
\multicolumn{2}{l}{\bf Fine-grained Detection} & & \\
\quad mT5$_{xxl}$ & 71.80 & 70.71 & 63.89 \\
\quad mT5$_{xxl}$ + \text{Exp} & 72.63 & 71.51 & 66.24 \\
\quad mT5$_{xxl}$ + \text{NLI} & 73.53 & 72.59 & 66.78 \\
\quad mT5$_{xxl}$ + \text{NLI} + \text{Exp} & \bf 74.27 & \bf  73.34 & \bf  67.52 \\
\bottomrule
\end{tabular}
}
\caption{The experiment results on supervised headline hallucination detection. The ``Coarse-grained Detection'' methods directly predict a coarse-grained label (``Support'', ``Neutral'', ``Contradict'') and are evaluated by the metric ``Accuracy'' and ``Weighted-F1''. The ``Fine-grained Detection'' methods predict a set of fine-grained labels (evaluated by ``Example-F1'') and we map them back to a single coarse-grained label. Please refer to Section~\ref{subsec:sft_exp_settings} for more details.}
\label{table:supervised_main_results}
\vspace*{-1.0em}
\end{table}

\subsection{Experiment Results}\label{subsec:sft_exp_results}

Table~\ref{table:supervised_main_results} presents our experiment results.
First, we observe that pre-training on NLI datasets (before the in-domain fine-tuning) can consistently enhance the model performance.
This improvement could stem from the shared characteristics between the headline hallucination detection task and the natural language inference task, both aiming to assess text grounding capability.
Second, we find that incorporating natural language explanations into models with the encoder-decoder architecture can significantly boost the model performance.
Similar observations are found in prior studies~\cite{Narang2020WT5TT,Shen2023WhyIT} and here our experiments verify the same phenomenon holds for multi-label fine-grained hallucination detection.

Besides incorporating NLI based pretraining and utilizing natural language explanation, we notice that for coarse-grained detection, it is generally preferable to initially train models for fine-grained prediction and then map fine-grained classes to coarse-grained ones.
Also, we want to stress that even the best performing method, mT5$_{xxl}$ + NLI + Exp, with 13B parameters and trained with in-domain data, still only achieves about 74\% detection accuracy.
This indicates the challenging nature of our MFHHD dataset, leaving plenty of room for future research improvements.
Finally, we show that the model trained on our benchmark can generalize to more hallucination detection datasets in Appendix~\ref{app:exp_true}.
%!TEX root = main.tex
% UTF-8 encoding

\section{Few-shot Fine-grained Headline Hallucination Detection}\label{sec:few-shot-exp}
We introduce various few-shot methods that leverages large language models for detecting fine-grained headline hallucinations in this section.

% Figure: llm based method
\begin{figure}[!t]
    \centering
    % \vspace{-ex}
    \includegraphics[width=0.91\linewidth]{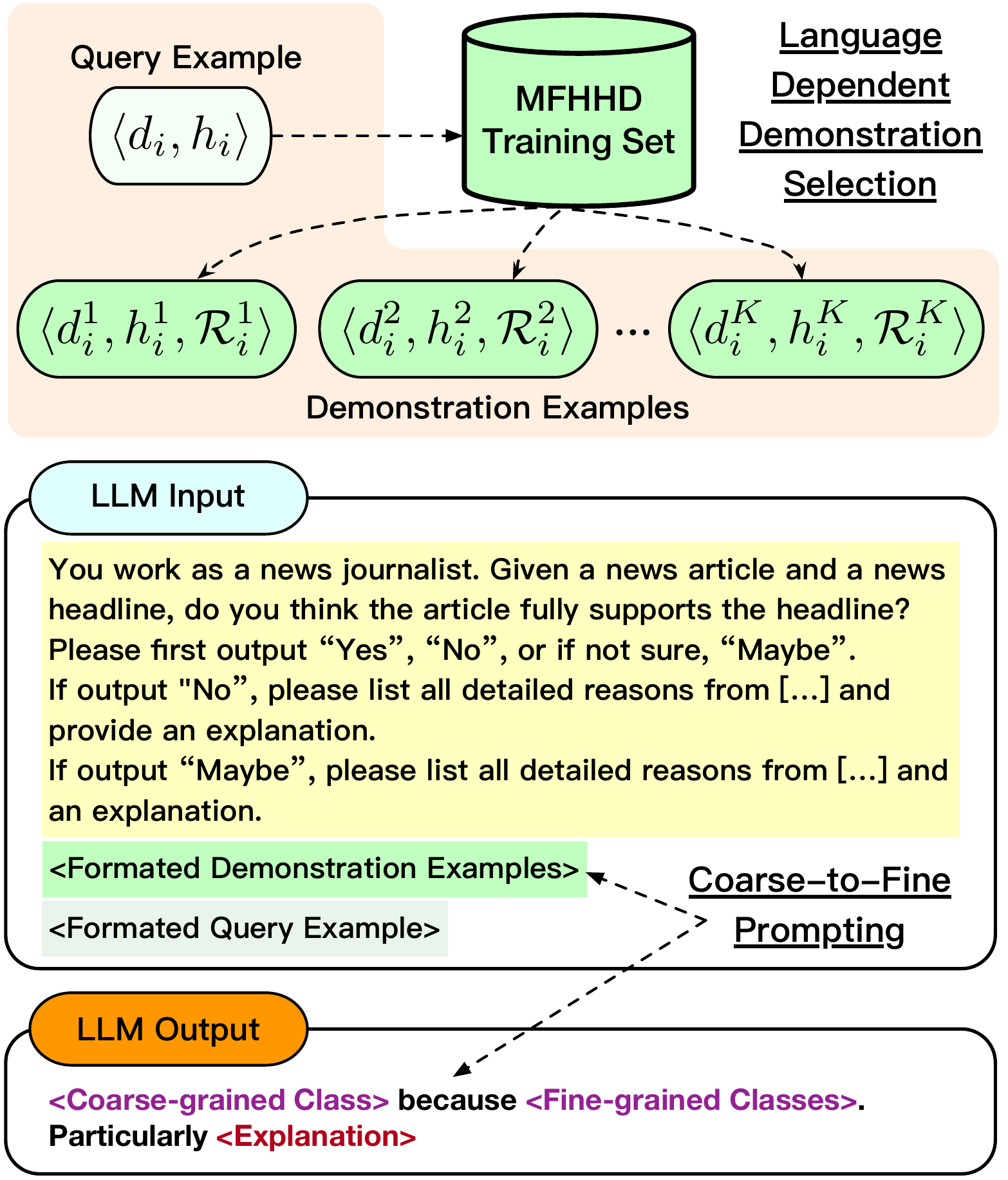}
    \vspace{-0.3ex}
    \caption{Detecting fine-grained headline hallucinations using LLM with language dependent demonstration selection and coarse-to-fine prompting.
    }
    \vspace{-2.5ex}
    \label{fig:icl_prompting}
\end{figure}

\subsection{LLM In-Context Learning (ICL)}
Recent studies have demonstrated that Large language models (LLMs) can quickly adapt to various tasks by learning only on a few demonstration examples in context~\cite{Brown2020LanguageMA,Wei2022ChainOT,chatgpt}.
Specifically, given a test example $\langle d_i, h_i \rangle$ and a set of $K$ demonstrations $\{ \langle d_i^{j}, h_i^{j}, \mathcal{R}_i^{j} \rangle | _{j=1}^{K}\}$ for the hallucination detection task, we will first format them using a template and then prompt the LLM to decode an output sequence that corresponds to the final prediction (c.f. Appendix~\ref{app:prompt_formats} for prompt template examples).

%% Table: Few-shot results
\begin{table*}[!tbp]
  \centering
  \resizebox{0.96\linewidth}{!}{
  \begin{tabular}{lc|ccc|ccc|ccc}
  \toprule
  \multirow{2}{*}{\textbf{Backbone}} & \multirow{2}{*}{\textbf{Methods}} & \multicolumn{3}{c}{\bfseries 1-shot} & \multicolumn{3}{c}{\bfseries 3-shot} & \multicolumn{3}{c}{\bfseries 5-shot} \\
  \cmidrule(lr){3-5} \cmidrule(lr){6-8} \cmidrule(lr){9-11}
   &  & Accuracy & Weighted-F1 & Example-F1 & Accuracy & Weighted-F1 & Example-F1 & Accuracy & Weighted-F1 & Example-F1 \\
  \midrule
  \multirow{4}{*}{PaLM2-L} 
  & LI-FG & 63.24 & 55.00 & 60.32 & 64.41 & 57.05 & 60.97 & 64.77 & 57.90 & 61.58  \\
  & LD-FG & 65.38 & 55.49 & 61.41 & 64.63 & 57.78 & 61.53 & 64.73 & 57.80 & 61.98 \\
  & LI-C2FG & 65.07 &	60.59 &	61.29 &		65.55 &	61.21 &	61.81 &		66.69 &	62.25 &	62.05 \\
  & LD-C2FG & \bf 65.67 & \bf	60.82 &	\bf 62.15 &		\bf 66.96 & \bf	63.01 & \bf	61.99 &	\bf	67.03 & \bf	62.90 & \bf	62.14 \\
  \midrule
  \multirow{4}{*}{PaLM2-M} 
  & LI-FG & 48.34 &	38.85 &	42.87 &		47.76 &	45.81 &	56.96 &		64.99 &	58.75 &	60.01 \\
  & LD-FG & 49.81 &	46.06 &	43.27 &		54.19 &	52.84 &	59.45 &		64.87 &	58.35 &	60.07 \\
  & LI-C2FG & 49.32 &	44.61 &	45.65 &		60.67 &	56.64 &	58.40 &		66.65 &	63.88 &	60.44 \\
  & LD-C2FG & \bf 56.63 &	\bf 46.73 &	\bf 47.71 &	\bf	64.88 &	\bf 62.01 &	\bf 59.65 &	\bf	67.21 & \bf	64.18 &	\bf 61.52 \\
  \midrule
  \multirow{4}{*}{PaLM2-S} 
  & LI-FG & 17.63 &	24.58 &	32.08 &		54.52 &	45.54 &	51.05 &		64.76 &	61.57 &	56.40 \\
  & LD-FG & 35.73 &	37.52 &	36.86 &		54.33 &	47.86 &	54.59 &		64.96 &	61.39 &	56.25 \\
  & LI-C2FG & 35.60 &	36.82 &	39.45 &		56.07 &	44.37 &	55.16 &		66.32 &	63.02 &	61.51 \\
  & LD-C2FG & \bf 45.00 &	\bf 47.26 &	\bf 49.54 &	\bf	57.31 & \bf	50.54 & \bf	58.68 &	\bf	66.49 &	\bf 63.21 & \bf	61.63 \\
  \bottomrule
  \end{tabular}
  }
  \caption{The experiment results on few-shot fine-grained headline hallucination detection. Prefixes ``LI'' and ``LD'' in method names stand for \underline{L}anguage-\underline{I}ndependent and \underline{L}anguage-\underline{D}ependent variants, respectively. We run each methods five times and report the averaged metrics.}
  \label{tab:few_shot_exp_results}
  \vspace{-0.5ex}
\end{table*}

%% Table: Ablation on demo languages
\begin{table}[!tbp]
    \resizebox{0.93\linewidth}{!}{
    \begin{tabular}{l|ccccc|c}   \toprule
    \bf Accuracy & EN & ES & DE & FR & PT & Avg.  \\ 
    \midrule
EN & \bf 80.70 & 51.40 & \bf 53.40 & 60.40 & 62.60 & 64.87 \\
ES & 78.40 & \bf 54.40 & 50.40 & 60.40 & 63.20 & 64.20 \\
DE & 80.20 & 52.80 & 51.40 & 60.40 & 64.60 & \bf 65.37 \\
FR & 80.10 & 52.80 & 51.40 & \bf 60.80 & \bf 65.20 & 65.07 \\
PT & 79.80 & 53.20 & 54.40 & 60.40 & 64.60 & \bf 65.37 \\
   \midrule 
   \bf Weighted-F1 & EN & ES & DE & FR & PT & Avg.  \\ 
    \midrule
EN & \bf 78.22 & 44.36 & \bf 45.05 & 52.21 & 56.31 & \bf 58.56 \\
ES & 77.24 & \bf 44.79 & 38.22 & 52.36 & 55.69 & 57.06 \\
DE & 78.78 & 42.57 & 43.26 & 53.46 & 55.39 & 58.14 \\
FR & 77.86 & 44.42 & 42.50 & \bf 54.24 & 56.22 & 58.29 \\
PT & 77.99 & 42.94 & 40.07 & 53.22 & \bf 57.33 & 57.72 \\
   \midrule
   \bf Example-F1 & EN & ES & DE & FR & PT & Avg.  \\ 
    \midrule
EN & \bf 75.63 & 49.90 & \bf 49.90 & 56.29 & 60.75 & 61.43 \\
ES & 75.30 & \bf 50.74 & 48.05 & 55.41 & 60.92 & 61.07 \\
DE & 74.99 & 49.37 & 48.15 & 55.70 & 61.31 & 60.75 \\
FR & 75.58 & 49.25 & 49.03 & \bf 56.79 & 60.91 & 61.21 \\
PT & 75.57 & 50.21 & 48.98 & 56.03 & \bf 61.64 & \bf 61.48 \\
   \bottomrule 
    \end{tabular}
    }
    \caption{Performance of PaLM2-L with 5-shot demonstration examples from various languages. We highlight the best demonstration example language (indicated by each row) for every test example language (indicated by each column).}
    \vspace{-1.5ex}
    \label{tab:few_shot_lang_specific}
\end{table}

\subsection{Language Dependent Demonstration Selection and Coarse-to-Fine Prompting}
In this work, we explore various LLMs to address the following research question: ``How to prompt LLMs for best few-shot fine-grained hallucination detection performance in a multilingual context?''.
We introduce two simple yet effective techniques to achieve this objective, illustrated in Figure~\ref{fig:icl_prompting}.

First, instead of using a fixed set of demonstrations for all test examples, we propose to select a dynamic set of demonstrations based on the language of the test example.
Different from most previous retrieval-based ICL studies~\cite{Luo2024IncontextLW}, this method does not reply on an external retrieval model and thus has a better application scope.
Second, we note that those fine-grained hallucination classes are interrelated rather than isolated.
For example, an instance cannot have both a ``Neutral'' and a ``Contradict'' subclass at the same time.
Given this hierarchical organization of hallucination labels, we present a coarse-to-fine prompting approach.
In this method, we prompt the LLM to produce an initial coarse-grained hallucination prediction, followed by more specific fine-grained predictions along with a natural language explanation.
We conduct extensive experiments to evaluate these two techniques with various LLMs.

\subsection{Experiment Settings}
For main experiments, we test PaLM2-S, PaLM2-M and PaLM2-L~\cite{anil2023palm} under 1-shot, 3-shot, and 5-shot settings.
All demonstrations are selected from the MFHHD training set.
We employ the same set of metrics as described in Section~\ref{subsec:sft_exp_settings} to compare the following methods:
(1) \textbf{LI-FG} uses a fixed set of demonstrations and directly prompts the LLM to output a set of fine-grained hallucination classes; 
(2) \textbf{LD-FG} selects language dependent demonstrations before prompting the LLM for direct fine-grained hallucination classes predictions;
(3) \textbf{LI-C2FG} adopts the coarse-to-fine prompting technique with a fixed set of demonstrations; and
(4) \textbf{LD-C2FG} combines both the coarse-to-fine prompting and language dependent demonstration selection techniques for fine-grained hallucination detection.

For all tested methods, we incorporate natural language explanations in a predict-then-explain pipeline~\cite{Lampinen2022CanLM} which outputs the explanations after the prediction and empirically works better than the Chain-of-Thought prompting~\cite{Wei2022ChainOT}.
Furthermore, we prompt LLM to generate 4 predictions and use self-consistency~\cite{Wang2022SelfConsistencyIC} to aggregate them into the final prediction. 
Lastly, to reduce the randomness in LLM calls, we create 5 different groups of demonstrations for each $k$-shot setting, and report the average performance over 5 runs.
Appendix~\ref{app:icl_exp_details} provides more implementation details and hyper-parameter settings.

\subsection{Experiment Results}

\noindent \textbf{Overall Results.}
Table~\ref{tab:few_shot_exp_results} exhibits the main experiment results.
First, we notice that increasing the number of demonstrations generally helps the model performance and has the most pronounced affects for small and medium sized LLMs.
Second, comparing LD-FG with LI-FG and LI-C2FG with LD-C2FG reveals that language dependent demonstration selection indeed helps us to more accurately identify fine-grained hallucination classes. 
Third, we compare those coarse-to-fine prompting methods with the direct fine-grained prediction methods, and observe that the initial predicted coarse-grained class does guide the LLM for better fine-grained predictions.
Finally, we note that even the best performing method (LD-C2FG with 5-shot PaLM2-L) still lags behind most supervised models with fewer parameters (c.f. Table~\ref{table:supervised_main_results}).
One reason could be the demonstrations in the prompt are not enough to fully convey the nuanced hallucination error type definitions.
We encourage more future studies to fill this performance gap between the supervised and few-shot methods.

\smallskip
\noindent \textbf{Effect of Demonstration Languages.}
We continue to evaluate how the language of demonstration examples affects the LLM hallucination detection performance.
Specifically, we prompt the LLM with demonstrations of the same language and evaluate the prediction quality for each test example language.
We report the 5-shot PaLM2-L performance (with coarse-to-fine prompting technique) in Table~\ref{tab:few_shot_lang_specific}.
First, we notice that forcing all demonstrations to have the same language will lead to worse performance (compared to 5-shot PaLM2-L LI-C2FG results in Table~\ref{tab:few_shot_exp_results}).
Second, we observe that LLM generally performs better on those test examples that have the same language as its input demonstrations.
The only exception is for German examples, it is better to prompt LLM with English demonstrations, which is somewhat understandable considering both languages share some grammatical similarities due to their common Germanic roots.
This observation further explains and verifies the effectiveness of our language dependent demonstration selection strategy.

\smallskip
\noindent \textbf{Effect of Prompting Methods.}
We continue to evaluate two additional variants of prompting methods: (1) Chain-of-Thought (CoT) which outputs the explanation before the final prediction and (2) Fine-to-Coarse (F2CG) that first outputs the fine-grained hallucination labels followed by a coarse-grained class.
Results are shown in Table~\ref{table:few_shot_prompt_methods}.
First, we notice that for both language dependent and independent methods, CoT performs worse than the predict-then-explain pipeline.
We hypothesize one reason could be the sparsity of explanation for non-hallucinated examples.
Namely, if a headline is fully supported by the article, the raters will not provide any explanation and the CoT will have to directly make the prediction, a behavior inconsistent with the hallucination cases.
Second, we find that coarse-to-fine prompting works significantly better than the fine-to-coarse prompting.
This is probably because coarse-grained hallucination prediction has less mistakes then the fine-grained prediction.
Therefore, it is better to condition on a more confident (i.e., coarse-grained) prediction for generating a less confident (i.e., fine-grained) prediction instead of the other way around.

\smallskip
\noindent \textbf{Generalization to more LLMs.}
We test how our proposed prompting methods generalize to more LLMs including ChatGPT~\cite{chatgpt} and GPT4~\cite{gpt4}\footnote{\small More experiment details are described in Appendix~\ref{app:gpt_exp_details}.}.
Results are exhibited in Table~\ref{table:openai_model_results}.
We can see that prompting with ChatGPT generally performs worse than PaLM2-L while GPT4 outperforms the PaLM2-L on fine-grained hallucination detection.
Furthermore, the experiment results still align with our prior findings that using language-dependent demonstrations and conducting coarse-to-fine prompting can consistently yield performance enhancements.

%% Table: Ablation on prompting methods
\begin{table}[!t]
\centering
\scalebox{0.72}{
\begin{tabular}{lcc|c}
\toprule
\bf Methods & \bf Accuracy & \bf Weighted-F1 & \bf Example-F1  \\
\midrule
LI-FG & 64.77 & 57.90 & 61.58 \\
LI-FG + CoT & 58.87 & 53.26 & 54.35  \\
LI-C2FG & \bf 66.69 & \bf 62.25 & \bf 62.05 \\
LI-F2CG & 61.57 & 56.96 & 58.05 \\
\midrule
LD-FG & 64.73 & 57.80 & 61.98 \\
LD-FG + CoT & 59.02 & 53.26 & 54.35 \\
LD-C2FG & \bf 67.03 & \bf 62.90 & \bf 62.14 \\
LD-F2CG & 63.37 & 58.04 & 61.22 \\
\bottomrule
\end{tabular}
}
\vspace*{-0.2cm}
\caption{Performance of PaLM2-L using 5-shot demonstration examples with different prompting methods.}
\label{table:few_shot_prompt_methods}
\vspace*{-1.0em}
\end{table}

\begin{table}[!t]
\centering
\scalebox{0.76}{
\begin{tabular}{lcc|c}
\toprule
\bf Backbone & \bf Accuracy & \bf Weighted-F1 & \bf Example-F1  \\
\midrule
\multicolumn{2}{l}{\bf ChatGPT} & & \\
\quad LI-FG & 43.40 & 47.40 & 36.52 \\
\quad LD-FG & 53.00 & 51.73 & 48.20 \\
\quad LD-C2FG & \bf 54.83 & \bf 53.53 & \bf 50.88 \\
\midrule
\multicolumn{2}{l}{\bf GPT4} & & \\
\quad LI-FG & 62.78 & 59.89 & 60.36 \\
\quad LD-FG & 63.54 & 63.20 & 63.24 \\
\quad LD-C2FG & \bf 65.88 & \bf 64.05 & \bf 64.59 \\
\bottomrule
\end{tabular}
}
% \vspace*{-0.2cm}
\caption{The experiment results of ChatGPT and GPT-4 on 1-shot fine-grained headline hallucination detection with different prompting methods.}
\label{table:openai_model_results}
\vspace*{-1.0em}
\end{table}
%!TEX root = main.tex
% UTF-8 encoding

\section{Related Work}\label{sec:related_work}

\noindent \textbf{Hallucination Detection.}
Hallucination, one long-standing issue for many natural language generation models, refers to the scenario where the generated content being nonsensical or inconsistent with the provided source content~\cite{Ji2022SurveyOH, Zhang2023SirensSI, Tonmoy2024ACS}.
Plenty of studies have been proposed to mitigate the hallucination issue by cleaning model training data~\cite{Nan2021EntitylevelFC, Goyal2021AnnotatingAM, Aharoni2022mFACEMS}, modifying model learning objectives~\cite{Stiennon2020LearningTS, nan2021improving}, designing better decoding algorithms~\cite{Sridhar2022ImprovedBS, Qiu2023ThinkWY}, and building specialized models to postprocess/filter generated contents~\cite{Cao2020FactualEC,Chen2021ImprovingFI,Shen2023WhyIT}.
At a high level, our study falls into the last category and focuses on multilingual fine-grained hallucination detection in the news domain.

\smallskip
\noindent \textbf{Fine-Grained Hallucination Evaluation.}
A few studies are proposed to evaluate hallucination error details for different downstream applications.
For text summarization, \citet{Goyal2020EvaluatingFI} proposes to categorize hallucination errors at the level of dependency arcs and \citet{Pagnoni2021UnderstandingFI} defines a hallucination typology based on frame semantics and linguistic discourse theory.
For text simplification, \citet{Devaraj2022EvaluatingFI} introduces a hallucination taxonomy based on the edit nature of simplification.
At the same time, \citet{Dziri2022OnTO} leverages the Verbal Response Modes to define hallucination errors in knowledge-grounded conversational models.
More recently, \citet{Mishra2024FinegrainedHD} proposes a hallucination taxonomy for open-ended LM generation without pre-determined grounding text.
Despite some promising results, these approaches either assume one example can only have one fine-grained hallucination label or test only on English examples.

\smallskip
\noindent \textbf{Multilingual Summarization Faithfulness.}
We can view the news headline generation as a special type of summarization and thus our study is also related to the research about improving faithfulness of multilingual summarization systems.
\citet{Aharoni2022mFACEMS} proposes to train a coarse-grained entailment model based on multilingual natural language inference datasets (e.g., XNLI~\cite{Conneau2018XNLIEC}, XTREME~\cite{Hu2020XTREMEAM}) and adopt it to filter unfaithful summaries in the training set.
\citet{Qiu2023DetectingAM} introduces a multilingual faithfulness evaluation metric by aggregating four English faithfulness metrics with a machine translator. 
Different from these studies, our work does not rely on a translation system and focuses more on identifying fine-grained hallucination types.

% \smallskip
% \noindent \textbf{LLM in context learning }

%!TEX root = main.tex
% UTF-8 encoding

\section{Conclusions and Future Work}\label{sec:conclusion}

This study explores multilingual fine-grained headline hallucination detection and introduces the MFHHD dataset --- a collection of over 11,000 expert-annotated examples across 5 languages with natural language explanations.
Through extensive experiments, we discover that supervised models gain from pre-training on NLI datasets and the integration of explanations into their outputs.
Additionally, LLM few-shot learners show improved performance when utilizing language-dependent demonstration selection and adopting a coarse-to-fine prompting strategy.
Interesting future research directions include (1) employing parameter-efficient tuning techniques to directly train LLMs on the MFHHD dataset, (2) annotating some fine-grained hallucination classes at the span level, and (3) expanding the MFHHD dataset to incorporate more languages and explore multi-document hallucination detection scenarios.

\section*{Limitations}
In this work, our primary goal is to identify the news headline hallucinations and thus define all fine-grained hallucination classes for the news domain applications.
We recognize that some of these fine-grained definitions will be too restricted or too lenient for other domains' applications.
How to effectively transfer the knowledge and signals in our MFHHD dataset to more general domain use cases would be an important research problem.
Furthermore, for the few-shot detection setting, we mostly test those proprietary LLMs as they demonstrate the strongest in-context learning capability.
Future work could explore whether and how various open-sourced LLMs can benefit most from our proposed prompting techniques.

\section*{Ethics Statement}
This work adheres to high ethical standards in its research methodology and execution. We obtain the multilingual, fine-grained news headline hallucination detection dataset through a meticulous annotation process, ensuring the dataset quality. 
By addressing the issue of hallucination in automated news headline generation across multiple languages, the study contributes positively to the integrity and accuracy of news dissemination.

\bibliography{main}

\begin{thebibliography}{45}
\expandafter\ifx\csname natexlab\endcsname\relax\def\natexlab#1{#1}\fi

\bibitem[{Aharoni et~al.(2022)Aharoni, Narayan, Maynez, Herzig, Clark, and Lapata}]{Aharoni2022mFACEMS}
Roee Aharoni, Shashi Narayan, Joshua Maynez, Jonathan Herzig, Elizabeth Clark, and Mirella Lapata. 2022.
\newblock \href {https://api.semanticscholar.org/CorpusID:263616472} {mface: Multilingual summarization with factual consistency evaluation}.
\newblock In \emph{ACL}.

\bibitem[{Anil et~al.(2023)Anil, Dai, Firat, Johnson, Lepikhin, Passos, Shakeri, Taropa, Bailey, Chen et~al.}]{anil2023palm}
Rohan Anil, Andrew~M Dai, Orhan Firat, Melvin Johnson, Dmitry Lepikhin, Alexandre Passos, Siamak Shakeri, Emanuel Taropa, Paige Bailey, Zhifeng Chen, et~al. 2023.
\newblock Palm 2 technical report.
\newblock \emph{arXiv preprint arXiv:2305.10403}.

\bibitem[{Brown et~al.(2020)Brown, Mann, Ryder, Subbiah, Kaplan, Dhariwal, Neelakantan, Shyam, Sastry, Askell, Agarwal, Herbert-Voss, Krueger, Henighan, Child, Ramesh, Ziegler, Wu, Winter, Hesse, Chen, Sigler, Litwin, Gray, Chess, Clark, Berner, McCandlish, Radford, Sutskever, and Amodei}]{Brown2020LanguageMA}
Tom~B. Brown, Benjamin Mann, Nick Ryder, Melanie Subbiah, Jared Kaplan, Prafulla Dhariwal, Arvind Neelakantan, Pranav Shyam, Girish Sastry, Amanda Askell, Sandhini Agarwal, Ariel Herbert-Voss, Gretchen Krueger, T.~J. Henighan, Rewon Child, Aditya Ramesh, Daniel~M. Ziegler, Jeff Wu, Clemens Winter, Christopher Hesse, Mark Chen, Eric Sigler, Mateusz Litwin, Scott Gray, Benjamin Chess, Jack Clark, Christopher Berner, Sam McCandlish, Alec Radford, Ilya Sutskever, and Dario Amodei. 2020.
\newblock \href {https://api.semanticscholar.org/CorpusID:218971783} {Language models are few-shot learners}.
\newblock \emph{NeurIPS}, abs/2005.14165.

\bibitem[{Cai et~al.(2023)Cai, Song, Cho, Wang, Wang, Yu, Liu, and Yu}]{Cai2023GeneratingUN}
Pengshan Cai, Kaiqiang Song, Sangwoo Cho, Hongwei Wang, Xiaoyang Wang, Hong Yu, Fei Liu, and Dong Yu. 2023.
\newblock \href {https://api.semanticscholar.org/CorpusID:259370694} {Generating user-engaging news headlines}.
\newblock In \emph{ACL}.

\bibitem[{Cao et~al.(2020)Cao, Dong, Wu, and Cheung}]{Cao2020FactualEC}
Mengyao Cao, Yue Dong, Jiapeng Wu, and Jackie Chi~Kit Cheung. 2020.
\newblock \href {https://api.semanticscholar.org/CorpusID:224706057} {Factual error correction for abstractive summarization models}.
\newblock \emph{EMNLP}, abs/2010.08712.

\bibitem[{Chen et~al.(2021)Chen, Zhang, Sone, and Roth}]{Chen2021ImprovingFI}
Sihao Chen, Fan Zhang, Kazoo Sone, and Dan Roth. 2021.
\newblock \href {https://api.semanticscholar.org/CorpusID:233296487} {Improving faithfulness in abstractive summarization with contrast candidate generation and selection}.
\newblock In \emph{NAACL}.

\bibitem[{Conneau et~al.(2018)Conneau, Lample, Rinott, Williams, Bowman, Schwenk, and Stoyanov}]{Conneau2018XNLIEC}
Alexis Conneau, Guillaume Lample, Ruty Rinott, Adina Williams, Samuel~R. Bowman, Holger Schwenk, and Veselin Stoyanov. 2018.
\newblock \href {https://api.semanticscholar.org/CorpusID:52271711} {Xnli: Evaluating cross-lingual sentence representations}.
\newblock In \emph{EMNLP}.

\bibitem[{Devaraj et~al.(2022)Devaraj, Sheffield, Wallace, and Li}]{Devaraj2022EvaluatingFI}
Ashwin Devaraj, William Sheffield, Byron~C. Wallace, and Junyi~Jessy Li. 2022.
\newblock \href {https://api.semanticscholar.org/CorpusID:248218448} {Evaluating factuality in text simplification}.
\newblock \emph{ACL}, 2022:7331--7345.

\bibitem[{Devlin et~al.(2019)Devlin, Chang, Lee, and Toutanova}]{Devlin2019BERTPO}
Jacob Devlin, Ming-Wei Chang, Kenton Lee, and Kristina Toutanova. 2019.
\newblock \href {https://api.semanticscholar.org/CorpusID:52967399} {Bert: Pre-training of deep bidirectional transformers for language understanding}.
\newblock In \emph{NAACL}.

\bibitem[{Ding et~al.(2023)Ding, Smith-Renner, Zhang, Tetreault, and Jaimes}]{Ding2023HarnessingTP}
Zijian Ding, Alison Smith-Renner, Wenjuan Zhang, Joel~R. Tetreault, and Alejandro Jaimes. 2023.
\newblock \href {https://api.semanticscholar.org/CorpusID:264172468} {Harnessing the power of llms: Evaluating human-ai text co-creation through the lens of news headline generation}.
\newblock In \emph{EMNLP}.

\bibitem[{Dziri et~al.(2022)Dziri, Milton, Yu, Zaiane, and Reddy}]{Dziri2022OnTO}
Nouha Dziri, Sivan Milton, Mo~Yu, Osmar~R Zaiane, and Siva Reddy. 2022.
\newblock \href {https://api.semanticscholar.org/CorpusID:248227301} {On the origin of hallucinations in conversational models: Is it the datasets or the models?}
\newblock In \emph{NAACL}.

\bibitem[{Goyal and Durrett(2020)}]{Goyal2020EvaluatingFI}
Tanya Goyal and Greg Durrett. 2020.
\newblock \href {https://api.semanticscholar.org/CorpusID:222291532} {Evaluating factuality in generation with dependency-level entailment}.
\newblock In \emph{ACL Findings}.

\bibitem[{Goyal and Durrett(2021)}]{Goyal2021AnnotatingAM}
Tanya Goyal and Greg Durrett. 2021.
\newblock \href {https://api.semanticscholar.org/CorpusID:233204406} {Annotating and modeling fine-grained factuality in summarization}.
\newblock In \emph{NAACL}.

\bibitem[{Gu et~al.(2020)Gu, Mao, Han, Liu, Yu, Wu, Yu, Finnie, Zhai, and Zukoski}]{Gu2020GeneratingRH}
Xiaotao Gu, Yuning Mao, Jiawei Han, Jialu Liu, Hongkun Yu, You Wu, Cong Yu, Daniel Finnie, Jiaqi Zhai, and Nicholas Zukoski. 2020.
\newblock \href {https://api.semanticscholar.org/CorpusID:210919915} {Generating representative headlines for news stories}.
\newblock \emph{WebConf}.

\bibitem[{He et~al.(2023)He, Gao, and Chen}]{He2023DeBERTaV3ID}
Pengcheng He, Jianfeng Gao, and Weizhu Chen. 2023.
\newblock \href {https://api.semanticscholar.org/CorpusID:244346093} {Debertav3: Improving deberta using electra-style pre-training with gradient-disentangled embedding sharing}.
\newblock \emph{ICLR}, abs/2111.09543.

\bibitem[{Honovich et~al.(2022)Honovich, Aharoni, Herzig, Taitelbaum, Kukliansy, Cohen, Scialom, Szpektor, Hassidim, and Matias}]{Honovich2022TRUERF}
Or~Honovich, Roee Aharoni, Jonathan Herzig, Hagai Taitelbaum, Doron Kukliansy, Vered Cohen, Thomas Scialom, Idan Szpektor, Avinatan Hassidim, and Yossi Matias. 2022.
\newblock \href {https://api.semanticscholar.org/CorpusID:247694170} {True: Re-evaluating factual consistency evaluation}.
\newblock In \emph{NAACL}.

\bibitem[{Hu et~al.(2020)Hu, Ruder, Siddhant, Neubig, Firat, and Johnson}]{Hu2020XTREMEAM}
Junjie Hu, Sebastian Ruder, Aditya Siddhant, Graham Neubig, Orhan Firat, and Melvin Johnson. 2020.
\newblock \href {https://api.semanticscholar.org/CorpusID:214641214} {Xtreme: A massively multilingual multi-task benchmark for evaluating cross-lingual generalization}.
\newblock \emph{ICML}, abs/2003.11080.

\bibitem[{Ji et~al.(2022)Ji, Lee, Frieske, Yu, Su, Xu, Ishii, Bang, Dai, Madotto, and Fung}]{Ji2022SurveyOH}
Ziwei Ji, Nayeon Lee, Rita Frieske, Tiezheng Yu, Dan Su, Yan Xu, Etsuko Ishii, Yejin Bang, Wenliang Dai, Andrea Madotto, and Pascale Fung. 2022.
\newblock \href {https://api.semanticscholar.org/CorpusID:246652372} {Survey of hallucination in natural language generation}.
\newblock \emph{ACM Computing Surveys}, 55:1 -- 38.

\bibitem[{Lampinen et~al.(2022)Lampinen, Dasgupta, Chan, Matthewson, Tessler, Creswell, McClelland, Wang, and Hill}]{Lampinen2022CanLM}
Andrew~Kyle Lampinen, Ishita Dasgupta, Stephanie C.~Y. Chan, Kory Matthewson, Michael~Henry Tessler, Antonia Creswell, James~L. McClelland, Jane~X. Wang, and Felix Hill. 2022.
\newblock \href {https://api.semanticscholar.org/CorpusID:247957917} {Can language models learn from explanations in context?}
\newblock \emph{EMNLP}, abs/2204.02329.

\bibitem[{Liu et~al.(2022)Liu, Swayamdipta, Smith, and Choi}]{Liu2022WANLIWA}
Alisa Liu, Swabha Swayamdipta, Noah~A. Smith, and Yejin Choi. 2022.
\newblock \href {https://api.semanticscholar.org/CorpusID:246016339} {Wanli: Worker and ai collaboration for natural language inference dataset creation}.
\newblock In \emph{EMNLP}.

\bibitem[{Luo et~al.(2024)Luo, Xu, Liu, Pasupat, and Kazemi}]{Luo2024IncontextLW}
Man Luo, Xin Xu, Yue Liu, Panupong Pasupat, and Mehran Kazemi. 2024.
\newblock \href {https://api.semanticscholar.org/CorpusID:267069067} {In-context learning with retrieved demonstrations for language models: A survey}.

\bibitem[{Mishra et~al.(2024)Mishra, Asai, Balachandran, Wang, Neubig, Tsvetkov, and Hajishirzi}]{Mishra2024FinegrainedHD}
Abhika Mishra, Akari Asai, Vidhisha Balachandran, Yizhong Wang, Graham Neubig, Yulia Tsvetkov, and Hannaneh Hajishirzi. 2024.
\newblock \href {https://api.semanticscholar.org/CorpusID:266999558} {Fine-grained hallucination detection and editing for language models}.

\bibitem[{Nan et~al.(2021{\natexlab{a}})Nan, Nallapati, Wang, dos Santos, Zhu, Zhang, McKeown, and Xiang}]{Nan2021EntitylevelFC}
Feng Nan, Ramesh Nallapati, Zhiguo Wang, C{\'i}cero~Nogueira dos Santos, Henghui Zhu, Dejiao Zhang, Kathleen McKeown, and Bing Xiang. 2021{\natexlab{a}}.
\newblock \href {https://api.semanticscholar.org/CorpusID:231951460} {Entity-level factual consistency of abstractive text summarization}.
\newblock In \emph{Conference of the European Chapter of the Association for Computational Linguistics}.

\bibitem[{Nan et~al.(2021{\natexlab{b}})Nan, Santos, Zhu, Ng, McKeown, Nallapati, Zhang, Wang, Arnold, and Xiang}]{nan2021improving}
Feng Nan, Cicero Nogueira~dos Santos, Henghui Zhu, Patrick Ng, Kathleen McKeown, Ramesh Nallapati, Dejiao Zhang, Zhiguo Wang, Andrew~O Arnold, and Bing Xiang. 2021{\natexlab{b}}.
\newblock Improving factual consistency of abstractive summarization via question answering.
\newblock \emph{arXiv preprint arXiv:2105.04623}.

\bibitem[{Narang et~al.(2020)Narang, Raffel, Lee, Roberts, Fiedel, and Malkan}]{Narang2020WT5TT}
Sharan Narang, Colin Raffel, Katherine Lee, Adam Roberts, Noah Fiedel, and Karishma Malkan. 2020.
\newblock \href {https://api.semanticscholar.org/CorpusID:216867225} {Wt5?! training text-to-text models to explain their predictions}.
\newblock \emph{ArXiv}, abs/2004.14546.

\bibitem[{Nie et~al.(2020)Nie, Williams, Dinan, Bansal, Weston, and Kiela}]{nie2019adversarial}
Yixin Nie, Adina Williams, Emily Dinan, Mohit Bansal, Jason Weston, and Douwe Kiela. 2020.
\newblock Adversarial nli: A new benchmark for natural language understanding.
\newblock In \emph{ACL}. Association for Computational Linguistics.

\bibitem[{OpenAI(2022)}]{chatgpt}
OpenAI. 2022.
\newblock Chatgpt.

\bibitem[{OpenAI(2023)}]{gpt4}
OpenAI. 2023.
\newblock Gpt-4 technical report.

\bibitem[{Pagnoni et~al.(2021)Pagnoni, Balachandran, and Tsvetkov}]{Pagnoni2021UnderstandingFI}
Artidoro Pagnoni, Vidhisha Balachandran, and Yulia Tsvetkov. 2021.
\newblock \href {https://api.semanticscholar.org/CorpusID:233407441} {Understanding factuality in abstractive summarization with frank: A benchmark for factuality metrics}.
\newblock \emph{NAACL}, abs/2104.13346.

\bibitem[{Perez et~al.(2021)Perez, Kiela, and Cho}]{Perez2021TrueFL}
Ethan Perez, Douwe Kiela, and Kyunghyun Cho. 2021.
\newblock \href {https://api.semanticscholar.org/CorpusID:235166749} {True few-shot learning with language models}.
\newblock \emph{NeurIPS}, abs/2105.11447.

\bibitem[{Prabhu et~al.(2018)Prabhu, Kag, Harsola, Agrawal, and Varma}]{prabhu2018parabel}
Yashoteja Prabhu, Anil Kag, Shrutendra Harsola, Rahul Agrawal, and Manik Varma. 2018.
\newblock Parabel: Partitioned label trees for extreme classification with application to dynamic search advertising.
\newblock In \emph{WebConf}, pages 993--1002.

\bibitem[{Qiu et~al.(2023{\natexlab{a}})Qiu, Embar, Cohen, and Han}]{Qiu2023ThinkWY}
Yifu Qiu, Varun Embar, Shay~B. Cohen, and Benjamin Han. 2023{\natexlab{a}}.
\newblock \href {https://api.semanticscholar.org/CorpusID:265221182} {Think while you write: Hypothesis verification promotes faithful knowledge-to-text generation}.
\newblock \emph{ArXiv}, abs/2311.09467.

\bibitem[{Qiu et~al.(2023{\natexlab{b}})Qiu, Ziser, Korhonen, Ponti, and Cohen}]{Qiu2023DetectingAM}
Yifu Qiu, Yftah Ziser, Anna Korhonen, E.~Ponti, and Shay~B. Cohen. 2023{\natexlab{b}}.
\newblock \href {https://api.semanticscholar.org/CorpusID:258841008} {Detecting and mitigating hallucinations in multilingual summarisation}.
\newblock \emph{EMNLP}, abs/2305.13632.

\bibitem[{Roberts et~al.(2022)Roberts, Chung, Levskaya, Mishra, Bradbury, Andor, Narang, Lester, Gaffney, Mohiuddin, Hawthorne, Lewkowycz, Salcianu, van Zee, Austin, Goodman, Soares, Hu, Tsvyashchenko, Chowdhery, Bastings, Bulian, Garcia, Ni, Chen, Kenealy, Clark, Lee, Garrette, Lee-Thorp, Raffel, Shazeer, Ritter, Bosma, Passos, Maitin-Shepard, Fiedel, Omernick, Saeta, Sepassi, Spiridonov, Newlan, and Gesmundo}]{roberts2022t5x}
Adam Roberts, Hyung~Won Chung, Anselm Levskaya, Gaurav Mishra, James Bradbury, Daniel Andor, Sharan Narang, Brian Lester, Colin Gaffney, Afroz Mohiuddin, Curtis Hawthorne, Aitor Lewkowycz, Alex Salcianu, Marc van Zee, Jacob Austin, Sebastian Goodman, Livio~Baldini Soares, Haitang Hu, Sasha Tsvyashchenko, Aakanksha Chowdhery, Jasmijn Bastings, Jannis Bulian, Xavier Garcia, Jianmo Ni, Andrew Chen, Kathleen Kenealy, Jonathan~H. Clark, Stephan Lee, Dan Garrette, James Lee-Thorp, Colin Raffel, Noam Shazeer, Marvin Ritter, Maarten Bosma, Alexandre Passos, Jeremy Maitin-Shepard, Noah Fiedel, Mark Omernick, Brennan Saeta, Ryan Sepassi, Alexander Spiridonov, Joshua Newlan, and Andrea Gesmundo. 2022.
\newblock \href {https://arxiv.org/abs/2203.17189} {Scaling up models and data with $\texttt{t5x}$ and $\texttt{seqio}$}.
\newblock \emph{arXiv preprint arXiv:2203.17189}.

\bibitem[{Shen et~al.(2023)Shen, Liu, Finnie, Rahmati, Bendersky, and Najork}]{Shen2023WhyIT}
Jiaming Shen, Jialu Liu, Daniel Finnie, Negar~Asgharipour Rahmati, Michael Bendersky, and Marc Najork. 2023.
\newblock \href {https://api.semanticscholar.org/CorpusID:256827429} {“why is this misleading?”: Detecting news headline hallucinations with explanations}.
\newblock \emph{WebConf}.

\bibitem[{Shen et~al.(2021)Shen, Qiu, Meng, Shang, Ren, and Han}]{Shen2021TaxoClassHM}
Jiaming Shen, Wenda Qiu, Yu~Meng, Jingbo Shang, Xiang Ren, and Jiawei Han. 2021.
\newblock \href {https://api.semanticscholar.org/CorpusID:235097263} {Taxoclass: Hierarchical multi-label text classification using only class names}.
\newblock In \emph{NAACL}.

\bibitem[{Sridhar and Visser(2022)}]{Sridhar2022ImprovedBS}
Arvind~Krishna Sridhar and Erik Visser. 2022.
\newblock Improved beam search for hallucination mitigation in abstractive summarization.
\newblock \emph{arXiv preprint arXiv:2212.02712}.

\bibitem[{Stiennon et~al.(2020)Stiennon, Long, Wu, Ziegler, Lowe, Voss, Radford, Amodei, and Christiano}]{Stiennon2020LearningTS}
Nisan Stiennon, Ouyang Long, Jeffrey Wu, Daniel~M. Ziegler, Ryan Lowe, Chelsea Voss, Alec Radford, Dario Amodei, and Paul~Francis Christiano. 2020.
\newblock \href {https://api.semanticscholar.org/CorpusID:263874153} {Learning to summarize with human feedback}.
\newblock In \emph{NeurIPS}.

\bibitem[{Tonmoy et~al.(2024)Tonmoy, Zaman, Jain, Rani, Rawte, Chadha, and Das}]{Tonmoy2024ACS}
S.~M Towhidul~Islam Tonmoy, S~M~Mehedi Zaman, Vinija Jain, Anku Rani, Vipula Rawte, Aman Chadha, and Amitava Das. 2024.
\newblock \href {https://api.semanticscholar.org/CorpusID:266725532} {A comprehensive survey of hallucination mitigation techniques in large language models}.
\newblock \emph{ArXiv}, abs/2401.01313.

\bibitem[{Wang et~al.(2022)Wang, Wei, Schuurmans, Le, hsin Chi, and Zhou}]{Wang2022SelfConsistencyIC}
Xuezhi Wang, Jason Wei, Dale Schuurmans, Quoc Le, Ed~Huai hsin Chi, and Denny Zhou. 2022.
\newblock \href {https://api.semanticscholar.org/CorpusID:247595263} {Self-consistency improves chain of thought reasoning in language models}.
\newblock \emph{ICML}, abs/2203.11171.

\bibitem[{Wei et~al.(2022)Wei, Wang, Schuurmans, Bosma, hsin Chi, Xia, Le, and Zhou}]{Wei2022ChainOT}
Jason Wei, Xuezhi Wang, Dale Schuurmans, Maarten Bosma, Ed~Huai hsin Chi, F.~Xia, Quoc Le, and Denny Zhou. 2022.
\newblock \href {https://api.semanticscholar.org/CorpusID:246411621} {Chain of thought prompting elicits reasoning in large language models}.
\newblock \emph{NeurIPS}, abs/2201.11903.

\bibitem[{Williams et~al.(2018)Williams, Nangia, and Bowman}]{MNLI}
Adina Williams, Nikita Nangia, and Samuel Bowman. 2018.
\newblock \href {http://aclweb.org/anthology/N18-1101} {A broad-coverage challenge corpus for sentence understanding through inference}.
\newblock In \emph{NAACL}, pages 1112--1122. Association for Computational Linguistics.

\bibitem[{Wolf et~al.(2020)Wolf, Debut, Sanh, Chaumond, Delangue, Moi, Cistac, Rault, Louf, Funtowicz, and Brew}]{Wolf2020HuggingFacesTS}
Thomas Wolf, Lysandre Debut, Victor Sanh, Julien Chaumond, Clement Delangue, Anthony Moi, Pierric Cistac, Tim Rault, R{\'e}mi Louf, Morgan Funtowicz, and Jamie Brew. 2020.
\newblock \href {https://api.semanticscholar.org/CorpusID:208117506} {Huggingface's transformers: State-of-the-art natural language processing}.
\newblock \emph{EMNLP (Demo)}, abs/1910.03771.

\bibitem[{Xue et~al.(2020)Xue, Constant, Roberts, Kale, Al-Rfou, Siddhant, Barua, and Raffel}]{Xue2020mT5AM}
Linting Xue, Noah Constant, Adam Roberts, Mihir Kale, Rami Al-Rfou, Aditya Siddhant, Aditya Barua, and Colin Raffel. 2020.
\newblock \href {https://api.semanticscholar.org/CorpusID:225040574} {mt5: A massively multilingual pre-trained text-to-text transformer}.
\newblock In \emph{North American Chapter of the Association for Computational Linguistics}.

\bibitem[{Zhang et~al.(2023)Zhang, Li, Cui, Cai, Liu, Fu, Huang, Zhao, Zhang, Chen, Wang, Luu, Bi, Shi, and Shi}]{Zhang2023SirensSI}
Yue Zhang, Yafu Li, Leyang Cui, Deng Cai, Lemao Liu, Tingchen Fu, Xinting Huang, Enbo Zhao, Yu~Zhang, Yulong Chen, Longyue Wang, Anh~Tuan Luu, Wei Bi, Freda Shi, and Shuming Shi. 2023.
\newblock \href {https://api.semanticscholar.org/CorpusID:261530162} {Siren's song in the ai ocean: A survey on hallucination in large language models}.
\newblock \emph{ArXiv}, abs/2309.01219.

\end{thebibliography}

%!TEX root = main.tex
% UTF-8 encoding

\clearpage
\appendix
% \onecolumn

\section{Appendix}\label{app:supp_materials}

\subsection{Fine-grained Hallucination Types with Illustrative Examples}\label{app:demo_examples}

\begin{figure*}[!t]
    \centering
    % \vspace{-ex}
    \includegraphics[width=0.98\linewidth]{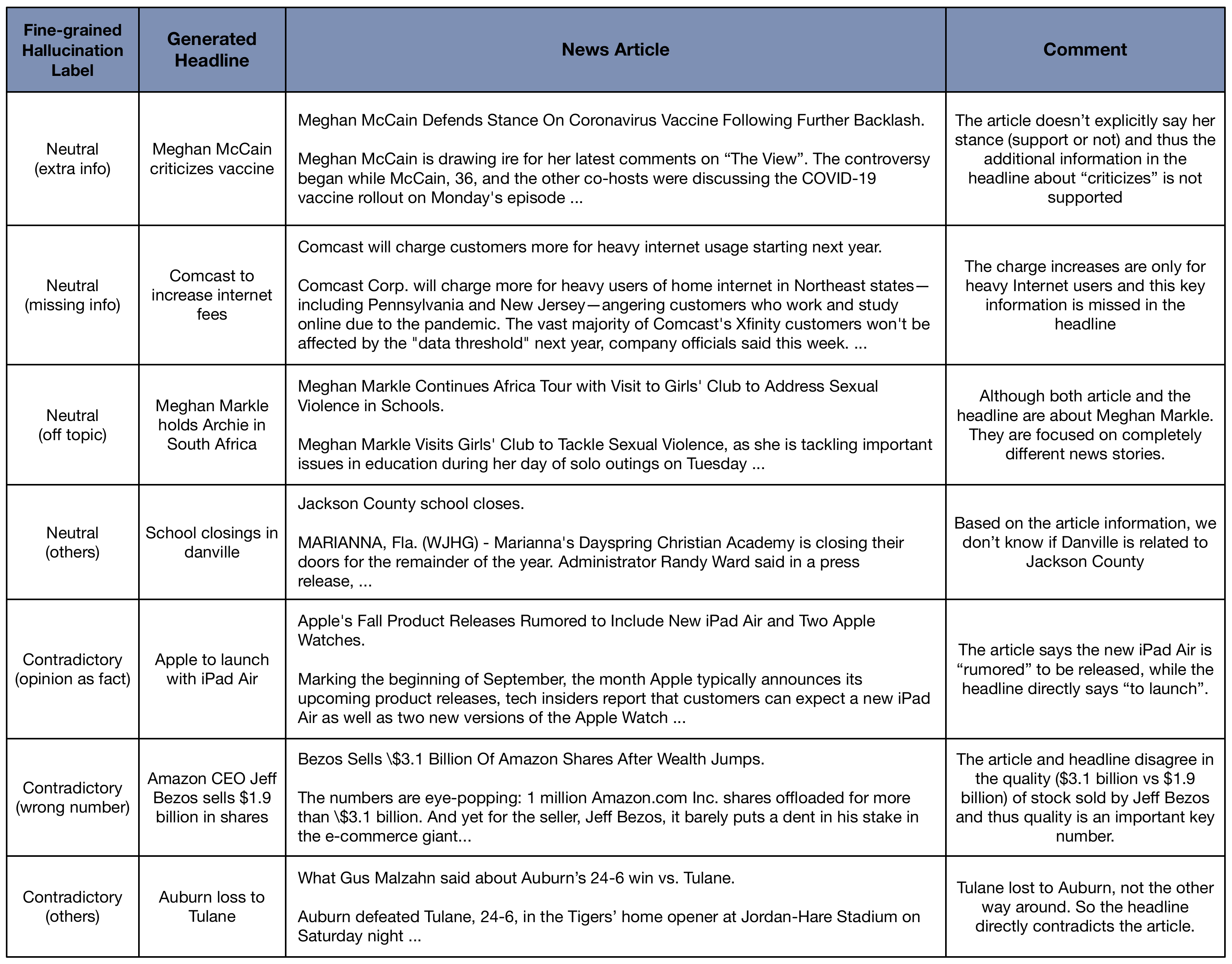}
    \vspace{-0.3ex}
    \caption{Fine-grained headline hallucination labels with illustrative examples.}
    % \vspace{-2ex}
    \label{fig:demo_examples}
\end{figure*}

We show one representative example for each fine-grained hallucination class in Figure~\ref{fig:demo_examples}.

\subsection{Discussions on Headline Sources}\label{app:label_source}
As discussion in main text, we collect our headlines from a system that utilizes NHNet for headline generation. 
We acknowledge that this is not a very diverse set of model generated headlines. 
Meanwhile, we want to emphasize that these headlines closely resemble real-world news headlines.

To further test if our model can generalize to other headline generation methods, we conduct an experiment where we first randomly select 25 examples from our MFHHD test split and use PaLM2-L and GPT4 to generate a headline that has the same fine-grained label(s) as the original headline. Then, we manually check that those generated headlines indeed have their corresponding labels (and if not, we will slightly modify their wordings to make them correctly labeled). 
Finally, we test our SFT model variant ``fine-grained mT5$_{xxl}$ + NLI + Exp'' (see Table~\ref{table:supervised_main_results} in the main text) and observe that it achieves accuracy = 0.64, example-F1 = 0.59. 
These results show that our model can generalize to more headline generation methods to some extent.
 
\subsection{Experiment Details on Supervised Hallucination Detection Methods}\label{app:sft_exp_details}
For mDeBERTa$_{{base}}$\footnote{\small \url{https://huggingface.co/microsoft/mdeberta-v3-base}} and mDeBERTa$_{{base}}$ + NLI\footnote{\small \url{https://huggingface.co/MoritzLaurer/mDeBERTa-v3-base-xnli-multilingual-nli-2mil7}}, we use their corresponding pre-trained checkpoints in the Huggingface Library.
We do parameter swamping on the learning rate in [5e-6, 1e-6, 5e-5, 1e-5] and perform three-fold cross validation on the training set. The final selected learning rate for mDeBERTa$_{{base}}$ is 1e-6 and learning rate for mDeBERTa$_{{base}}$ + NLI is 5e-6.
Finally, we train both models on a single A100-40GB with batch size 8 for 3 epochs.

For the remaining four mT5$_{xxl}$ based models, we implement them using the T5X library\footnote{\small \url{https://github.com/google-research/t5x}} with pre-trained mT5 checkpoints\footnote{\small \url{https://github.com/google-research/multilingual-t5}}.
due to computational constraints, we directly use their default hyper-parameters.
Specifically, we set the batch size to be 128, constant learning rate to be 1e-3, and the maximum output tokens to be 128. 
If the output sequence exceeds the length limit (e.g., having a long human written explanation in mT5$_{xxl}$ + Exp), we will simply truncate the output sequences to its first 128 tokens. 
We train all four models on TPU v3 for 10k steps with 1k warmup steps.

\subsection{Experiments on TRUE Benchmark}\label{app:exp_true}
We investigate if the SFT model trained on our benchmark can generalize to more hallucination detection datasets.
Specifically, we evaluate the model variant ``fine-grained mT5$_xxl$ + NLI + Exp'' (see Table~\ref{table:supervised_main_results} in the main text) on the TRUE benchmark~\cite{Honovich2022TRUERF}. 
Table~\ref{table:results_on_true} reports the experiment results. 
The Q2 and ANLI are two best performing methods in the original TRUE paper. We can see that the model trained on our MFHHD dataset has good zero-shot performance on the TRUE benchmark, which demonstrates the broad applicability of our dataset and the generalization ability of our method.

%% Table: TRUE results
\begin{table}[!t]
\centering
\scalebox{0.75}{
\begin{tabular}{lcc|c}
\toprule
\bf Dataset & \bf Q2 & \bf ANLI & \bf mT5$_{xxl}$ + \text{NLI} + \text{Exp}   \\
\midrule
MNBM        & 66.5 & 66.7 & \bf 70.44 \\
FRANK       & \bf 82.9 & 83.5 & 75.85 \\
QAGS        & \bf 78.3 & 75.3 & 72.76 \\
SummEval    & 77.3 & 72.9 & \bf 85.81 \\
FEVER       & 82.7 & \bf 90.2 & 88.24 \\
Vitamin-C   & 75.7 & 74.7 & \bf 84.71 \\
\midrule
Average & 77.23	& 77.22	& \bf 79.64 \\
\bottomrule
\end{tabular}
}
\caption{The experiment results on TRUE benchmark.}
\label{table:results_on_true}
\vspace*{-1.0em}
\end{table}

\subsection{Experiment Details on Few-shot Hallucination Detection Methods}\label{app:icl_exp_details}

\noindent \textbf{Demonstration Selection.}
We implement all of our $k$-shot experiments in the true few-shot setting~\cite{Perez2021TrueFL} for the multi-class classification problem. 
Specifically, we will sample $k$ demonstrations for each coarse-grained hallucination label and do not assume the presence of a large labeled development set of hyper-parameter tuning. 
Namely, we will have 3 demonstration examples for 1-shot setting, 9 demonstrations for 3-shot setting, and 15 demonstrations for 5-shot setting.
For the language-independent prompting methods, we use the same set of demonstrations for all test examples.
For the language-dependent prompting methods, we will sample the corresponding number of demonstrations for each language and dynamically choose the demonstration set based on the test example language.
Furthermore, to reduce the LLM call randomness, we repeat the above sampling procedure 5 times and report the averaged performance over these 5 independent runs.

\smallskip 
\noindent \textbf{Explanation Order.}
In our experiments, we test both the chain-of-thought (CoT) prompting~\cite{Wei2022ChainOT} which generates explanations \emph{before} making predictions and predict-then-explain prompting~\cite{Lampinen2022CanLM} which outputs explanations \emph{after} making predictions.
For our hallucination detection task, we observe that the predict-then-explain prompting consistently outperforms CoT prompting, particularly for small and medium sized LLMs.
Therefore, we choose to use the predict-then-explain prompting in this work.

We hypothesize that the ineffectiveness of CoT in our task comes from two aspects. 
First, we do NOT manually write any CoT demonstration and directly use the expert written explanations. 
These explanations may not be the best rationales for CoT and thus impair its performance. 
Second, we think there is an intrinsic difference between the hallucination detection task and the math reasoning task (e.g., GSM8K). The former one does not really require complicated multi-step reasoning and typically reaches the final decision in one or two steps. 
For example, if a rater witnesses an entity mismatch or ``rumor as fact" statement, he/she will directly label the headline as contradictory. This effectively shrinks the CoT improvement room.

\smallskip 
\noindent \textbf{Self Consistency.}
\citet{Wang2022SelfConsistencyIC} proposes the self-consistency method which samples multiple output sequences from the LLM and aggregates them into the final prediction.
In our experiments, we set the default temperature $t = 0.7$ and sample 4 decoded sequences from the LLM.
For each decoded sequence, we first parse it into a set of fine-grained labels.
Then, we select all fine-grained labels that appear in at least 2 decoded sequences as the final predicted hallucination labels.

\subsection{Prompt Templates}\label{app:prompt_formats}

\begin{lstlisting}[style=mystyle, caption={Prompt Template for Direct Fine-Grained Headline Hallucination Prediction}, label=lst:prompt, escapeinside={<@}{@>}]
You work as a news journalist. Given a news article and a news headline, you need to determine the relation between this article and the headline. Possible relations include: ["match", "incorrect_number", "opinion_as_fact", "direct_contradiction", "unsupported_additional_info", "miss_important_info", "off_topic", "neither_support_nor_contradiction"].

Please first output all possible relations and then provide an explanation.
See the following examples for references.

# demonstrations
Article: <@\textcolor{blue}{[article]}@>
Headline: <@\textcolor{blue}{[headline]}@>
Output: <@\textcolor{blue}{[class]}@>

Now, you are given a new news article and a news headline. Think step by step and then make the prediction.

# test examples
Article: <@\textcolor{blue}{[article]}@>
Headline: <@\textcolor{blue}{[headline]}@>
Output: <@\textcolor{red}{[class]}@>
\end{lstlisting}

\begin{lstlisting}[style=mystyle, caption={Prompt Template for Coarse-to-Fine Headline Hallucination Prediction}, label=lst:prompt, escapeinside={<@}{@>}]
You work as a news journalist. Given a news article and a news headline, do you think the article fully supports the headline? Please first output "Yes", "No", or if not sure, "Maybe".
If output "No", please list all detailed reasons from ["opinion_as_fact", "incorrect_number", "direct_contradiction"] and provide an explanation.
If output "Maybe", please list all detailed reasons from ["miss_important_info", "unsupported_additional_info", "off_topic", "neither_support_nor_contradiction"] and provide an explanation.
See the following examples for references.

# demonstrations
Article: <@\textcolor{blue}{[article]}@>
Headline: <@\textcolor{blue}{[headline]}@>
Output: <@\textcolor{blue}{[class]}@>

Now, you are given a new news article and a news headline. Think step by step and then make the prediction.

# test examples
Article: <@\textcolor{blue}{[article]}@>
Headline: <@\textcolor{blue}{[headline]}@>
Output: <@\textcolor{red}{[class]}@>
\end{lstlisting}

\subsection{Experiments on ChatGPT and GPT4}\label{app:gpt_exp_details}
We use the \texttt{gpt-3.5-turbo-0125} model for experimenting ChatGPT and adopt the \texttt{gpt-4-0125-preview} model for GPT4.
Due to the budget considerations, we only decode 1 generation from each model and remove the self consistency aggregation.
When experimenting coarse-to-fine prompting using ChatGPT, we observe that the decoded sequence occasionally fails to follow the ideal  output format (c.f. Figure~\ref{fig:icl_prompting}) and we will re-query the model using the JSON mode.

\end{document}